\documentclass[lettersize,journal]{IEEEtran}
\usepackage{amsmath,amsfonts}
\usepackage{algorithmic}
\usepackage{algorithm}
\usepackage{array}
\usepackage[caption=false,font=normalsize,labelfont=sf,textfont=sf]{subfig}
\usepackage{textcomp}
\usepackage{stfloats}
\usepackage{url}
\usepackage{verbatim}
\usepackage{graphicx}
\usepackage{cite}
\usepackage{bm}
\usepackage{amssymb}
\usepackage{xcolor}
\usepackage{subcaption}
\usepackage{tikz}
\usetikzlibrary{shapes.geometric, arrows.meta, positioning, calc}
\usepackage{hyperref}
\usepackage{soul}
\usepackage{orcidlink}
\hypersetup{colorlinks=true, citecolor=blue, linkcolor=red, urlcolor=blue}
\usepackage{multirow}
\usepackage[table]{xcolor}
\usepackage{xcolor}
\definecolor{darkGreen}{RGB}{0,100,0}
\begin{document}
\bstctlcite{IEEEexample:BSTcontrol}

\title{Temporal Learning for End-Effector Position Estimation under Aerodynamic Disturbances in Aerial Continuum Manipulation}

\author{Niloufar Amiri, Houman Masnavi, Farrokh Janabi-Sharifi% <-this % stops a space
\thanks{This work was supported by the Natural Sciences and Engineering Research Council of Canada (NSERC) under Grant 2023-05542 and by the National Research Council Canada (NRC) under Grant AI4L-128-1.}%

\thanks{Niloufar Amiri and Farrokh Janabi-Sharifi are with the Department of Mechanical, Industrial, and Mechatronics Engineering, Toronto Metropolitan University, Toronto, ON, Canada. Houman Masnavi is a Research Fellow at the University of Freiburg, Freiburg, Germany. E-mail: niloufar.amiri@torontomu.ca (corresponding author).}%
%

% <-this % stops a space
% \thanks{Manuscript received April 19, 2021; revised August 16, 2021.}
}

% The paper headers
% \markboth{Journal of \LaTeX\ Class Files,~Vol.~14, No.~8, August~2021}%
% {Shell \MakeLowercase{\textit{et al.}}: A Sample Article Using IEEEtran.cls for IEEE Journals}

% \IEEEpubid{0000--0000/00\$00.00~\copyright~2021 IEEE}
% Remember, if you use this you must call \IEEEpubidadjcol in the second
% column for its text to clear the IEEEpubid mark.

\maketitle

\begin{abstract}

This paper investigates temporal neural networks for \mbox{end-effector} position \mbox{estimation} of an aerial continuum manipulator (ACM) operating under aerodynamic effects induced by the unmanned aerial vehicle (UAV). An experimental dataset is collected under stationary (\mbox{rotor-off}) and \mbox{free-hovering} conditions across continuum robot (CR) configurations and UAV altitudes, providing \mbox{end-effector} position measurements with and without aerodynamic residuals. To establish a nominal framework, \mbox{strain-parameterized} kinematic models with progressively richer strain bases are evaluated to balance model complexity and prediction accuracy. The selected nominal model then serves as the baseline for 3D position residual estimation using a \mbox{closed-form} \mbox{continuous-time} (CfC) neural network, with a multilayer perceptron (MLP) and a gated recurrent unit (GRU) used for comparison. On unseen test experiments, the CfC achieves an RMSE of \(22.00\pm1.70~\mathrm{mm}\) over five random seeds, compared with \(36.38\pm3.58~\mathrm{mm}\) for the MLP and \(27.72\pm2.92~\mathrm{mm}\) for the GRU, corresponding to reductions of \(39.52\%\) and \(20.62\%\), respectively. These results demonstrate the effectiveness of \mbox{continuous-time} learning for \mbox{end-effector} position estimation under aerodynamic disturbances relative to static and \mbox{discrete-time} learning methods.

\end{abstract}


\section{Introduction}

The integration of soft and continuum manipulators into aerial robotic systems enables enhanced dexterity in elevated and confined environments and potentially safer physical interaction with surrounding objects \cite{jalali2022aerial}. These advantages, however, introduce additional modeling and estimation challenges due to the compliant and nonlinear behavior of aerial continuum manipulators (ACMs) \cite{amiri2025high,11598604}. In particular, their lightweight and deformable structures make them more susceptible than rigid aerial manipulators to aerodynamic disturbances induced by the unmanned aerial vehicle (UAV) \cite{khamseh2018aerial,ollero2021past}, leading to measurable deviations in the continuum robot (CR) configuration and \mbox{end-effector} position \cite{uthayasooriyan2026experimental}.

Accurate position estimation under these conditions requires both an appropriate nominal representation of the CR and a means of capturing the residual effects introduced by the induced airflow. Establishing the nominal response is itself nontrivial, since the accuracy of strain-parameterized continuum models depends on the spatial order used to approximate the strain distribution along the backbone. Increasing the strain basis order can capture more complex spatial variations, but also introduces additional generalized coordinates and model complexity \cite{renda2020geometric,boyer2020dynamics}. An appropriate nominal model should therefore provide sufficient accuracy without unnecessarily high-order parameterizations that yield only marginal improvement \cite{amiri2026strain}.

The \mbox{end-effector} position residual under aerodynamic disturbances presents an additional challenge because the downwash loading experienced by the CR varies with its configuration and UAV operating conditions, while the compliant response can depend on motion history through hysteresis and delayed effects. Consequently, the resulting position deviation may not be adequately represented by instantaneous system variables alone. Static neural models can approximate nonlinear input--output relationships but do not retain temporal information across sequential samples. Recurrent neural networks maintain a hidden memory of previous observations, while \mbox{closed-form} continuous-time networks additionally model the time-dependent evolution of this state, providing a representation more naturally aligned with continuously evolving physical dynamics \cite{lipton2015critical,hasani2022cfc}. These characteristics motivate the investigation of temporal neural networks for \mbox{end-effector} position estimation under \mbox{UAV-induced} aerodynamic uncertainty in ACMs.

\section{Related Work}

Disturbances in aerial manipulation are commonly treated as lumped uncertainties incorporating aerodynamic effects, platform vibration, modeling errors, and UAV--manipulator coupling. Model-based approaches, including nonlinear disturbance observers and adaptive methods, have been widely employed for their estimation and rejection \cite{liang2022adaptive,chen2022adaptive,liang2024observer}. However, these methods generally rely on sufficiently accurate system models and assumptions regarding the structure or bounds of unknown disturbances, which can be difficult to establish for highly coupled and nonlinear systems subject to complex aerodynamic interactions. Learning-based approaches have therefore been employed to approximate unmodeled dynamics and external disturbances using adaptive radial basis function (RBF) networks and deep neural networks \cite{fang2023robust,li2024adaptive,wu2024robust,wang2023millimeter}.

More specifically, aerodynamic disturbances have been investigated using analytical and data-driven approaches \cite{bauersfeld2024robotics,chen2021adaptive,wang2023neural,li2023nonlinear,kharitenko2025spatiotemporal}, although these studies primarily consider individual UAVs or aerodynamic interactions among multiple aerial vehicles. For ACMs, recent experiments have shown that rotor downwash can significantly affect CR deformation and \mbox{end-effector} position \cite{uthayasooriyan2026experimental,zhang2026motion}. In the most closely related study, Uthayasooriyan et al. \cite{uthayasooriyan2026experimental} developed a Gaussian process regression model guided by constant curvature (CC) kinematics for \mbox{steady-state} residual correction using a mechanically constrained multirotor. In contrast, this work considers a free-hovering UAV, establishes a strain-parameterized nominal model through a systematic study of spatial strain complexity, and estimates 3D position residuals using static, \mbox{discrete-time} recurrent, and \mbox{continuous-time} neural models across CR configurations and hovering altitudes. This formulation extends \mbox{steady-state} correction toward temporal residual estimation, explicitly examining the roles of temporal memory, \mbox{continuous-time} modeling, and nominal model complexity in uncertain position estimation.

The main contributions are:
\begin{itemize}

\item An experimental dataset characterizing ACM \mbox{end-effector} position under stationary and free-hovering UAV conditions across CR configurations, hovering altitudes, and UAV operating conditions.

\item A comparative evaluation of strain-parameterized kinematic models with increasing spatial strain basis order to establish a compact nominal representation for aerodynamic residual estimation.

\item A systematic comparison of MLP, GRU, and CfC networks for 3D \mbox{end-effector} position residual estimation under free-hovering UAV conditions, extending existing \mbox{steady-state}, CC-based correction by evaluating temporal memory and \mbox{continuous-time} modeling.

\end{itemize}

\section{Problem Formulation and Methodology}
\subsection{Nominal Strain-Parameterized Model}

Let $s\in[0,L]$ denote the arc-length coordinate along a CR, where $L$ is its
undeformed length. The configuration of a \mbox{cross-section} at $s$ relative to the
base frame is represented by
\begin{equation}
    \bm g(s)=
    \begin{bmatrix}
        \bm R(s) & \bm r(s)\\
        \bm 0_{1\times3} & 1
    \end{bmatrix}
    \in\mathrm{SE}(3),
    \label{eq:cr_configuration}
\end{equation}
where $\bm R(s)\in\mathrm{SO}(3)$ and $\bm r(s)\in\mathbb R^3$ denote the
\mbox{cross-section} orientation and position, respectively. Following strain-parameterized Cosserat rod theory
\cite{boyer2020dynamics,renda2020geometric}, the configuration evolves as
\begin{equation}
    \bm g'(s)=\bm g(s)\widehat{\bm\xi}(s),
    \qquad
    \widehat{\bm\xi}(s)=
    \begin{bmatrix}
        \operatorname{S}(\bm\kappa) & \bm\nu\\
        \bm 0_{1\times3} & 0
    \end{bmatrix},
    \label{eq:cr_kinematics}
\end{equation}
where $\operatorname{S}(\cdot)$ is the skew-symmetric operator and
$\bm\xi=[\bm\kappa^\top,\bm\nu^\top]^\top\in\mathbb R^6$ is the local strain
vector. Here, $\bm\kappa\in\mathbb R^3$ contains the bending and torsional
strains, while $\bm\nu\in\mathbb R^3$ contains the axial and shear strains.

The distributed strain field is approximated by
\begin{equation}
    \bm\xi(s,\bm q_s)
    =
    \bm\xi^\star+
    \bm B_m\!\left(\frac{s}{L}\right)\bm q_s,
    \label{eq:strain_basis}
\end{equation}
where $\bm\xi^\star$ is the reference strain,
$\bm q_s\in\mathbb R^{n_s}$ contains the generalized strain coordinates, and
$\bm B_m\in\mathbb R^{6\times n_s}$ is the spatial strain basis of order $m$.
The corresponding forward kinematics (FK) is
\begin{equation}
\begin{aligned}
    \bm g(s;\bm q_s)
    ={}& \bm g(0)\,
    \mathcal P\exp\!\Bigg(
    \int_0^s
    \widehat{\bm\xi}(\sigma,\bm q_s)\,d\sigma
    \Bigg),
\end{aligned}
    \label{eq:strain_fk}
\end{equation}
where $\mathcal P$ denotes the path-ordering operator. The kinematics are
numerically evaluated using Lie-group integration to preserve the
$\mathrm{SE}(3)$ structure \cite{renda2020geometric}.

\begin{figure*}[t]
    \centering
    \includegraphics[width=0.90\textwidth]{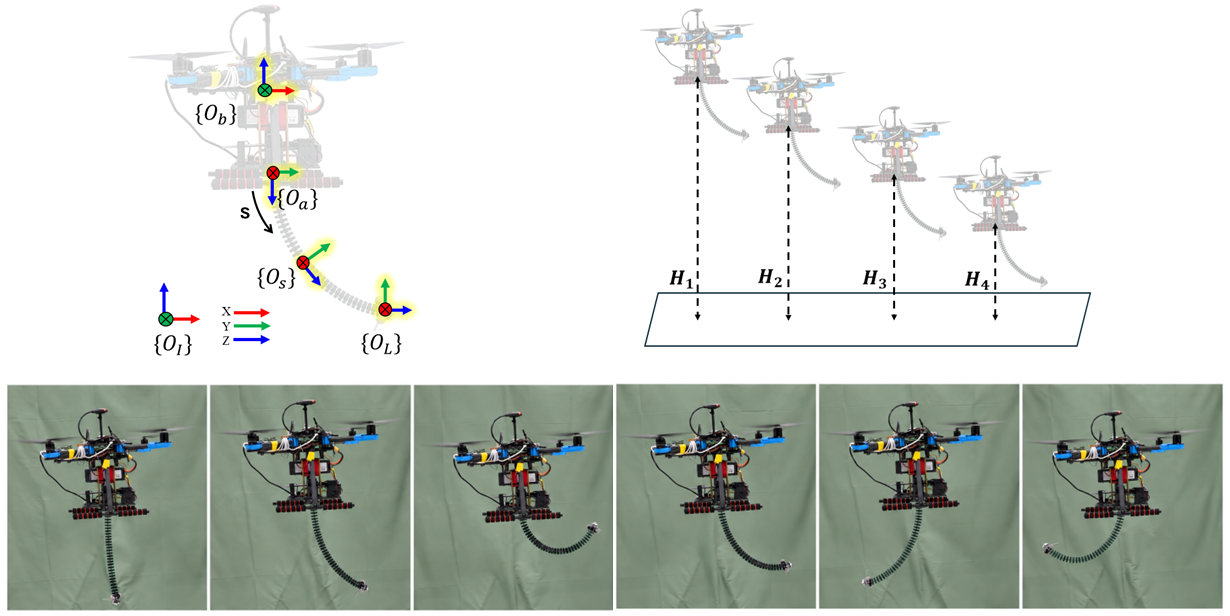}
    \caption{ACM frame attachment and operating conditions across different CR configurations.}
    \label{fig:intro}
\end{figure*}

For the tendon-driven CR considered in this work, shown in
Fig.~\ref{fig:intro}, the base frame is the attachment frame $O_a$, such that
$\bm g(0)=\bm I_4$. The straight, unsheared reference strain is written as
$\bm\xi^\star=[\bm 0_3^\top,\bm e_b^\top]^\top$, where $\bm e_b$ is the unit
vector along the undeformed backbone in the adopted local frame. Only the two
actuated bending strains are parameterized, while torsion, extension, and shear
remain fixed at their reference values. Shifted Legendre polynomials are employed
as spatial basis functions. Constant, linear, quadratic, and cubic orders,
corresponding to $m=0,1,2,3$, are evaluated to examine the tradeoff between
nominal prediction accuracy and spatial model complexity. Based on this
comparison, the linear basis is selected for the nominal model.

For the selected model, the two bending strain distributions are
\begin{equation}
\begin{aligned}
    \kappa_{1}(s) &=
    A_1\!\left[1+a_1P_1\!\left(\frac{s}{L}\right)\right],\\
    \kappa_{2}(s) &=
    A_2\!\left[1+a_2P_1\!\left(\frac{s}{L}\right)\right],
\end{aligned}
\label{eq:linear_strain_profiles}
\end{equation}
where $P_1(\eta)=2\eta-1$ is the first-order shifted Legendre polynomial.
Accordingly,
\begin{equation}
    \bm q_s=
    \begin{bmatrix}
        A_1 & a_1A_1 & A_2 & a_2A_2
    \end{bmatrix}^{\top},
    \label{eq:selected_qs}
\end{equation}
where $A_1$ and $A_2$ are bending amplitudes, while
$a_1$ and $a_2$ govern their spatial variation.

The bending amplitudes are related to the measured actuator coordinates
$\bm q_a=[q_1,q_2]^\top$ through a calibrated actuation mapping
\begin{equation}
    \begin{bmatrix}
        A_1\\
        A_2
    \end{bmatrix}
    =
    \bm\Phi(\bm q_a;\bm\theta_{\Phi}),
    \label{eq:actuator_amplitudes}
\end{equation}
where $\bm\Phi(\cdot)$ denotes the actuator-to-strain mapping and
$\bm\theta_{\Phi}$ collects its calibration parameters. The mapping is
calibrated separately from the spatial strain parameters. For the selected linear strain model, the remaining spatial
parameters are collected as
$\bm\theta_s=[a_1,a_2]^\top$.

The nominal \mbox{end-effector} position is obtained from the translational
component $\bm r_L=\bm r(L)$ of the tip configuration as
\begin{equation}
    \widehat{\bm p}^{\mathrm{nom}}
    =
    \bm C\bm r_L+\bm p_0,
    \label{eq:nominal_tip}
\end{equation}
where $\bm C\in\mathbb R^{3\times3}$ and $\bm p_0\in\mathbb R^3$ denote the
fixed coordinate registration and translation used to express the model
prediction in the experimental measurement frame.

After calibration of the actuator-to-strain mapping, the spatial parameters
are identified from the stationary baseline measurements by minimizing the
\mbox{end-effector} position error,
\begin{equation}
    \bm\theta_s^\star
    =
    \operatorname*{arg\,min}_{\bm\theta_s}
    \sum_{i=1}^{N_{\mathrm{stat}}}
    \left\|
        \bm p_i^{\mathrm{stat}}
        -
        \widehat{\bm p}^{\mathrm{nom}}
        (\bm q_{a,i};\bm\theta_s)
    \right\|_2^2,
    \label{eq:nominal_identification}
\end{equation}
where $N_{\mathrm{stat}}$ is the number of stationary measurements and
$\bm p_i^{\mathrm{stat}}\in\mathbb R^3$ is the corresponding measured
\mbox{end-effector} position. The resulting nominal model is fixed before the
subsequent aerodynamic residual learning process.

\subsection{UAV-Induced Residual Estimation}

The nominal model represents the CR response identified under stationary
(rotor-off) conditions. Under free-hovering conditions, the measured
\mbox{end-effector} position deviates from this nominal response due to
UAV-induced aerodynamic effects and associated operating conditions. To
estimate this deviation, the neural network input at sample $k$ is defined as
\begin{equation}
    \bm u_k
    =
    \begin{bmatrix}
        q_{1,k} &
        q_{2,k} &
        \dot q_{1,k} &
        h_k &
        \tau_k
    \end{bmatrix}^{\top},
    \label{eq:nn_input}
\end{equation}
where $q_{1,k}$ and $q_{2,k}$ are the actuator coordinates,
$\dot q_{1,k}$ is the reciprocating actuation rate, $h_k$ is the UAV
hovering altitude, and $\tau_k$ is the recorded throttle signal.

For the corresponding actuator configuration, the 3D position residual is
defined as
\begin{equation}
    \Delta\bm p_k
    =
    \bm p^{\mathrm{hov}}_k
    -
    \widehat{\bm p}^{\mathrm{nom}}_k,
    \label{eq:aero_residual}
\end{equation}
where $\bm p^{\mathrm{hov}}_k\in\mathbb R^{3}$ is the measured CR
\mbox{end-effector} position during free hovering and
$\widehat{\bm p}^{\mathrm{nom}}_k$ is the corresponding prediction of the nominal model. The residual is estimated as
\begin{equation}
    \widehat{\Delta\bm p}_k
    =
    \mathcal{F}(\bm u_k,\bm u_{k-1},\ldots),
    \label{eq:residual_general}
\end{equation}
where the dependence on preceding observations is determined by the selected
neural architecture. The corrected position estimate is then
\begin{equation}
    \widehat{\bm p}^{\mathrm{hov}}_k
    =
    \widehat{\bm p}^{\mathrm{nom}}_k
    +
    \widehat{\Delta\bm p}_k.
    \label{eq:total_tip_prediction}
\end{equation}

Because $\Delta\bm p_k$ is defined relative to an experimentally calibrated
nominal model, it can contain both UAV-induced position deviations and residual
nominal model mismatch. It is therefore interpreted as an effective
UAV-induced position residual rather than a direct measurement or estimate of
aerodynamic force.

\subsection{Comparative Neural Architectures}

Three neural architectures with different temporal modeling capabilities are
considered: an MLP, a GRU, and a CfC network. The MLP provides a memoryless
baseline, the GRU introduces discrete-time recurrent memory, and the CfC
employs a closed-form state transition derived from continuous-time neural
dynamics \cite{hasani2022cfc}.

For the recurrent models, consecutive measurements are arranged into finite
input sequences. For a recurrent architecture with sequence length $N_s$, a
sequence ending at sample $k$ is defined as
\begin{equation}
\bm U_k^{(N_s)}
=
\left[
\bm u_{k-N_s+1},\ldots,\bm u_k
\right].
\label{eq:input_sequence}
\end{equation}
The sequence length $N_s$ is selected separately for the GRU and CfC
architectures. Sequences are constructed independently within each experiment,
and a temporal gap exceeding a prescribed threshold $\Delta t_{\max}$
initiates a new sequence segment. The MLP, GRU, and CfC models are evaluated
at identical target timestamps to ensure a direct comparison.

\subsubsection{Multilayer Perceptron}

The MLP estimates the residual from the instantaneous input vector. For hidden
layer $\ell$,
\begin{equation}
\bm a_k^{(\ell)}
=
\rho\left(
\bm W^{(\ell)}
\bm a_k^{(\ell-1)}
+
\bm b^{(\ell)}
\right),
\qquad
\bm a_k^{(0)}=\bm u_k,
\label{eq:mlp_hidden}
\end{equation}
and the output is
\begin{equation}
\widehat{\Delta\bm p}_k
=
\bm W_o^{\mathrm{MLP}}
\bm a_k^{(L_m)}
+
\bm b_o^{\mathrm{MLP}}.
\label{eq:mlp_output}
\end{equation}
Here, $\bm W^{(\ell)}$ and $\bm b^{(\ell)}$ denote the trainable weight
matrix and bias vector of hidden layer $\ell$, respectively, while
$\bm W_o^{\mathrm{MLP}}$ and $\bm b_o^{\mathrm{MLP}}$ denote the
output-layer parameters. The function $\rho(\cdot)$ denotes the nonlinear
activation function, and $L_m$ is the number of hidden layers. The MLP
therefore implements the static mapping
\begin{equation}
\widehat{\Delta\bm p}_k
=
\mathcal{F}_{\mathrm{MLP}}(\bm u_k),
\label{eq:mlp_mapping}
\end{equation}
without an explicit recurrent state.

\subsubsection{Gated Recurrent Unit}

The GRU introduces a recurrent hidden state $\bm h_k$ to retain information
from preceding observations \cite{cho2014learning}. Its reset and update gates are defined as
\begin{align}
\bm r_k
&=
\sigma\left(
\bm W_{ir}\bm u_k+\bm b_{ir}
+
\bm W_{hr}\bm h_{k-1}+\bm b_{hr}
\right),\\
\bm z_k
&=
\sigma\left(
\bm W_{iz}\bm u_k+\bm b_{iz}
+
\bm W_{hz}\bm h_{k-1}+\bm b_{hz}
\right),
\label{eq:gru_gates}
\end{align}
with candidate hidden state
\begin{equation}
\begin{aligned}
\widetilde{\bm h}_k
=
\tanh\Big(
\bm W_{in}\bm u_k+\bm b_{in}
+\bm r_k\odot
(\bm W_{hn}\bm h_{k-1}+\bm b_{hn})
\Big).
\end{aligned}
\label{eq:gru_candidate}
\end{equation}
The recurrent state is updated according to
\begin{equation}
\bm h_k
=
(\bm 1-\bm z_k)\odot\widetilde{\bm h}_k
+
\bm z_k\odot\bm h_{k-1},
\label{eq:gru_state}
\end{equation}
and mapped to the residual estimate through
\begin{equation}
\widehat{\Delta\bm p}_k
=
\bm W_o^{\mathrm{GRU}}\bm h_k
+
\bm b_o^{\mathrm{GRU}}.
\label{eq:gru_output}
\end{equation}
Here, $\bm W_{i(\cdot)}$ and $\bm W_{h(\cdot)}$ denote the trainable input
and recurrent weight matrices, respectively, and $\bm b_{i(\cdot)}$ and
$\bm b_{h(\cdot)}$ are the corresponding bias vectors. The parameters
$\bm W_o^{\mathrm{GRU}}$ and $\bm b_o^{\mathrm{GRU}}$ define the output
mapping. The function $\sigma(\cdot)$ denotes the sigmoid function and
$\odot$ is the Hadamard product.

\subsubsection{Closed-Form Continuous-Time Network}

The CfC architecture employs a closed-form recurrent state transition derived
from continuous-time neural dynamics, avoiding numerical integration of the
underlying state evolution \cite{hasani2022cfc}. Defining the augmented input
to the recurrent transition as
\begin{equation}
\bm\eta_k
=
\begin{bmatrix}
\bm u_k^{\top} &
\bm h_{k-1}^{\top}
\end{bmatrix}^{\top}.
\label{eq:cfc_augmented_input}
\end{equation} The elapsed time between consecutive measurements is defined as $\Delta t_k = t_k-t_{k-1}$, where $t_k$ denotes the timestamp associated with sample $k$. A
time-dependent interpolation gate is then expressed as
\begin{equation}
\bm\alpha_k
=
\sigma\left(
\mathcal{N}_a(\bm\eta_k)\Delta t_k
+
\mathcal{N}_b(\bm\eta_k)
\right),
\label{eq:cfc_gate}
\end{equation}
and the recurrent state is updated according to
\begin{equation}
\begin{aligned}
\bm h_k
={}&
(\bm 1-\bm\alpha_k)
\odot\mathcal{N}_1(\bm\eta_k)
+
\bm\alpha_k
\odot\mathcal{N}_2(\bm\eta_k),
\end{aligned}
\label{eq:cfc_state}
\end{equation}
where $\mathcal{N}_1$, $\mathcal{N}_2$, $\mathcal{N}_a$, and
$\mathcal{N}_b$ denote trainable nonlinear mappings. The dependence on
$\Delta t_k$ allows the recurrent transition to account explicitly for
variations in the time interval between consecutive measurements. The
residual estimate is obtained as
\begin{equation}
\widehat{\Delta\bm p}_k
=
\bm W_o^{\mathrm{CfC}}\bm h_k
+
\bm b_o^{\mathrm{CfC}}.
\label{eq:cfc_output}
\end{equation}

\section{Experimental Setup and Dataset Generation}

Fig.~\ref{fig:intro} shows the experimental platform used for dataset
generation, consisting of a tendon-driven CR rigidly mounted on a UAV. The CR
workspace is partitioned into 12 slices defined by prescribed ranges of
$q_1$. For each experiment, $q_1$ is set to the midpoint of the corresponding
range and held constant, while $q_2$ undergoes reciprocating motion. Thus,
$\dot q_1=0$ during each experiment and is not included as a model input.
By symmetry, each experiment represents four workspace slices; consequently,
three experiments cover all 12 slices at a given hovering altitude.

The experiments are repeated at four hovering altitudes, resulting in 12
free-hovering experiments. Five additional baseline experiments are conducted
with inactive rotors to characterize the nominal CR behavior without
UAV-induced airflow. During each experiment, the CR is actuated sufficiently
slowly such that inertial effects are negligible and its motion can be treated
as kinetostatic. The 3D positions of the CR end-effector and UAV center of mass
are measured at \(10~\mathrm{Hz}\) using a Vicon motion capture system (Vicon
Motion Systems Ltd., UK), while UAV throttle and the two tendon-actuating servo
encoder measurements are recorded.

The complete dataset contains 28,411 samples, including 10,538 baseline samples and 17,873 samples collected under UAV operation. For residual estimation, Tests~13--16 (with payload) are excluded, and domain filtering of Tests~1--12 retains 13,922 free-hovering samples. The data are split at the experiment level, with Tests~1--9 forming the development dataset, Tests~10--12 reserved as the unseen dataset for final evaluation, and Tests~17--21 used exclusively for nominal model calibration.

\section{Results and Analysis}

\subsection{Nominal Model Evaluation}

\begin{figure*}[t]
    \centering
    \includegraphics[width=0.90\linewidth]{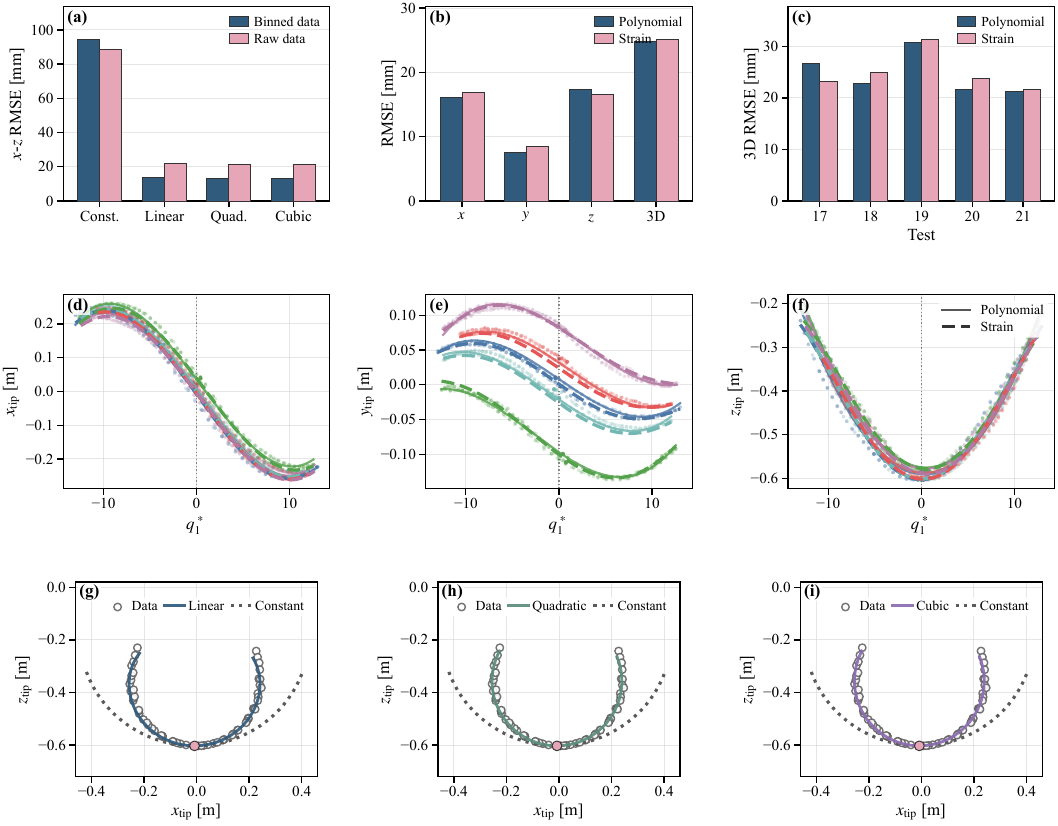}
    \caption{Nominal model evaluation. (a) \(x\)–\(z\) RMSE of strain models with different basis orders on Test~17 using binned and raw data. (b) RMSE of the third-order polynomial and selected linear strain models over the complete baseline dataset. (c) Per-test RMSE for Tests~17--21. (d)--(f) Measured and predicted \(x_{\mathrm{tip}}\), \(y_{\mathrm{tip}}\), and \(z_{\mathrm{tip}}\) versus centered \(q_1^*\) across the five baseline tests; colors distinguish tests, and solid/dashed curves denote polynomial/strain models. (g)--(i) Test~17 \(x\)–\(z\) trajectories comparing measured data and the constant strain model with the linear, quadratic, and cubic strain models, respectively.}
    \label{fig:FK}
\end{figure*}

The nominal CR kinematics is identified from the baseline dataset in two steps. First, strain-parameterized models with constant through cubic spatial bases are compared using Test~17 ($q_2=0$), where deformation occurs predominantly in the principal bending plane. Measurements with similar $q_1$ values are averaged to approximate the nominal static relationship and reduce variability due to reciprocating motion. As shown in Fig.~\ref{fig:FK}(a) and (b), the constant strain model yields a binned $x$--$z$ RMSE of $94.72~\mathrm{mm}$, whereas the linear basis reduces the error to $13.84~\mathrm{mm}$, an $85.4\%$ reduction. Quadratic and cubic bases provide only marginal further improvement, reaching approximately $13.03~\mathrm{mm}$. The linear basis is therefore selected as a compact representation of the dominant nonuniform deformation. These results indicate that the CC assumption, equivalent to constant strain in the present formulation, poorly represents the actual CR deformation.

The selected strain model is then evaluated over the complete baseline dataset against a purely data-driven third-order polynomial FK . As shown in Fig.~\ref{fig:FK}(c)--(g), the two models achieve comparable accuracy across the workspace. The polynomial model yields RMSEs of $(16.13,,7.53,,17.37)~\mathrm{mm}$ in $x$, $y$, and $z$, respectively, with a 3D RMSE of $24.87~\mathrm{mm}$, compared with $(16.94,,8.48,,16.60)~\mathrm{mm}$ and $25.19~\mathrm{mm}$ for the strain model. Despite similar accuracy, the strain model and actuation mapping together use 20 identified parameters, compared with 30 coefficients in the purely data-driven polynomial model, while preserving a physically structured representation of the distributed CR deformation rather than directly fitting the Cartesian tip coordinates.

\subsection{UAV-Induced Residual Characterization}

\begin{figure*}[t]
    \centering
    \includegraphics[width=0.90\linewidth]{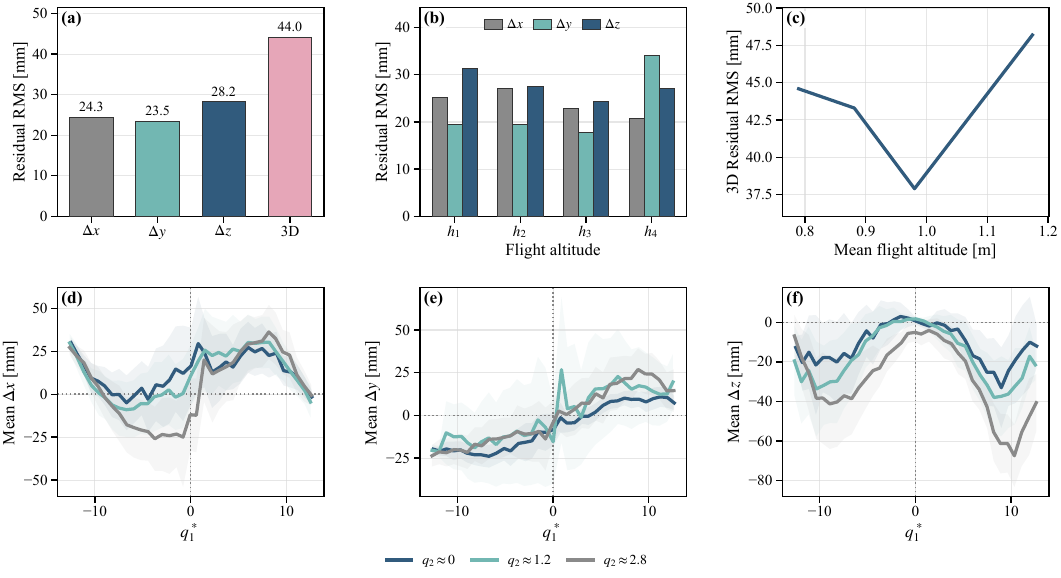}
    \caption{Aerodynamic residual characterization for free-hovering Tests~1--12. (a) Overall RMS of the Cartesian residual components $\Delta x$, $\Delta y$, and $\Delta z$, and the 3D residual. (b) Directional residual RMS at four flight altitudes $h_1$--$h_4$, ordered by increasing mean altitude. (c) 3D residual RMS versus mean flight altitude. (d)--(f) Mean $\Delta x$, $\Delta y$, and $\Delta z$ versus centered $q_1^*$ for the three configuration slices $q_2\approx0$, $q_2\approx1.2$, and $q_2\approx2.8$. Solid curves show binned mean values, with shaded regions indicating $\pm1$ standard deviation.}
    \label{fig:residual_characterization}
\end{figure*}

Before evaluating the residual estimators, the UAV-induced position residual is characterized across the CR workspace and hovering conditions to quantify its 3D magnitude and dependence on CR coordinates and UAV altitude. Across Tests~1--12, the residual has an overall 3D RMS of $44.05~\mathrm{mm}$, as shown in Fig.~\ref{fig:residual_characterization}(a). The residual varies with both flight altitude and CR configuration. Fig.~\ref{fig:residual_characterization}(b) and (c) show a nonmonotonic dependence on altitude, while Fig.~\ref{fig:residual_characterization}(d)--(f) reveal nonlinear variations with $q_1^*$ that also differ across the $q_2$ configuration slices. Residuals are present in all three Cartesian directions, with the largest RMS component along $z$, consistent with the primary direction of the rotor-induced downwash acting on the CR beneath the UAV.

\subsection{Learning-Based Residual Estimation}

The development dataset is utilized to train and compare the MLP, GRU, and CfC
estimators for 3D UAV-induced residual prediction. Hyperparameters are selected
using three-fold cross-validation, with each fold using one experiment from
each development altitude for validation and the remaining experiments for
training. Thus, all development altitudes are represented in both partitions.
Input and target standardization statistics are computed exclusively from each
training partition and the unseen dataset remains reserved for final
evaluation. Cross-validation selects sequence lengths of 25 and 50 samples for the GRU and
CfC, respectively, with 32 recurrent units and a learning rate of
$10^{-3}$ for both models. The MLP contains two hidden layers of 64 ReLU units
each. All models are trained using the Adam optimizer and mean squared error (MSE)
loss.

\begin{figure*}[t]
    \centering
    \includegraphics[width=0.90\linewidth]{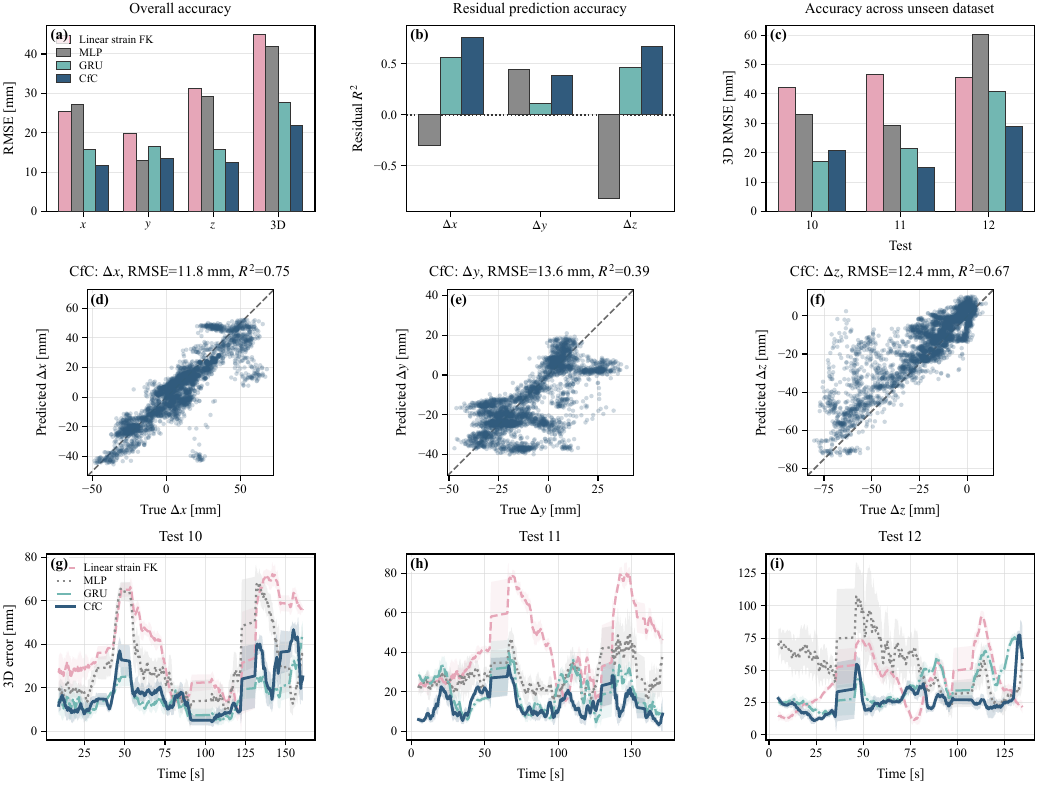}
    \caption{Residual estimators performance on unseen dataset.}
    \label{fig:NN_3D}
\end{figure*}

Final performance is evaluated on the unseen dataset, collected at a hovering
altitude excluded from neural network development, with the nominal linear
strain-parameterized model used as the uncompensated reference.
Fig.~\ref{fig:NN_3D} summarizes the prediction performance and shows the
estimated residuals and instantaneous position errors across the unseen
experiments. As shown in Fig.~\ref{fig:NN_3D}(a), the nominal model yields a
3D RMSE of $44.94~\mathrm{mm}$, which is reduced to $41.98~\mathrm{mm}$ by
the MLP. The temporal models provide substantially larger improvements,
reducing the 3D RMSE to $27.75~\mathrm{mm}$ for the GRU and
$21.87~\mathrm{mm}$ for the CfC. The directional errors show a similar
trend, with particularly notable improvements in the local $x$ and $z$
directions.

The residual $R^2$ values in Fig.~\ref{fig:NN_3D}(b) further demonstrate the
benefit of temporal modeling. The MLP yields negative $R^2$ values for
$\Delta x$ and $\Delta z$, indicating poor generalization of the memoryless
mapping for these residual components. In contrast, the GRU achieves positive
$R^2$ values in all three directions, while the CfC reaches approximately
$0.75$, $0.39$, and $0.67$ for $\Delta x$, $\Delta y$, and $\Delta z$,
respectively. The distribution of true versus predicted values in
Fig.~\ref{fig:NN_3D}(d)--(f) similarly shows that the CfC captures the dominant
residual trends, although greater dispersion remains in the $y$ direction,
which may reflect the more limited workspace variation along this dimension.
The time histories in Fig.~\ref{fig:NN_3D}(g)--(i) further show that the
temporal models generally maintain lower instantaneous 3D position errors
across the unseen experiments, although performance varies with operating
conditions and Test~12 remains the most challenging case. These
results show that incorporating temporal information substantially improves
residual estimation compared with a memoryless nonlinear mapping for the
present ACM system.

\subsection{Model Sensitivity and Robustness}

Input ablation and robustness analyses are conducted to evaluate the contribution of each input to prediction accuracy and the sensitivity of the learned estimators to stochastic training. For the input ablation, the selected
temporal models are retrained by individually removing $\dot q_1$, hovering
altitude, and rotor throttle while keeping the remaining model settings fixed.
The same three-fold cross-validation protocol used during model development is
applied to each ablation case to ensure a fair comparison. Robustness is
evaluated by independently training each final model configuration with five
random seeds and evaluating each model on the same unseen dataset,
without further hyperparameter adjustment.

Table~\ref{tab:input_ablation} summarizes the input ablation results using the
same cross-validation employed during model development. Removing
the actuation rate $\dot q_1$ increases the mean 3D RMSE by $11.34\%$ for the
CfC and $11.20\%$ for the GRU. This consistent degradation indicates that rate information captures recent CR motion and delayed residual effects beyond the instantaneous configuration, supporting its inclusion in the estimator input.

Hovering altitude also contributes to prediction accuracy. Removing altitude
increases the 3D RMSE by $18.89\%$ for the CfC and $7.52\%$ for the GRU.
For the CfC, the largest directional degradation occurs in the $y$
direction, where the RMSE increases from $18.27$ to $28.00~\mathrm{mm}$.
This behavior is consistent with the residual characterization, which shows
that the aerodynamic residual varies with hovering altitude without following
a simple monotonic trend. In contrast, removing rotor throttle slightly
reduces the validation RMSE by $2.05\%$ for the CfC and $3.60\%$ for the
GRU. Although rotor thrust directly influences the generated downwash, the UAV
operated near its available thrust capacity because of the limited payload
margin, resulting in a relatively narrow throttle range during the
experiments. Consequently, the throttle signal provided limited additional
predictive information in the present dataset.

\begin{table*}[t]
\centering
\caption{Input ablation analysis for the temporal estimators. The reported
$\Delta\mathrm{RMSE}$ is relative to the corresponding full-input model, with
positive values indicating increased error.}
\label{tab:input_ablation}

\begin{tabular*}{\textwidth}{@{\extracolsep{\fill}}llccccc@{}}
\hline
\textbf{Model} &
\textbf{Removed} &
\textbf{3D RMSE [mm]} &
\textbf{$\Delta\mathrm{RMSE}$ [\%]} &
\textbf{$x$ [mm]} &
\textbf{$y$ [mm]} &
\textbf{$z$ [mm]} \\
\hline

\multirow{4}{*}{CfC}
& None
& $29.62 \pm 11.59$
& --
& 17.48
& 18.27
& 13.47 \\

& $\dot{q}_1$
& $32.97 \pm 10.86$
& $+11.34$
& 22.68
& 15.91
& 16.84 \\

& Altitude
& $35.21 \pm 7.03$
& $+18.89$
& 16.63
& 28.00
& 11.37 \\

& Throttle
& $\mathbf{29.01 \pm 11.36}$
& $-2.05$
& 16.36
& 17.76
& 14.18 \\
\hline

\multirow{4}{*}{GRU}
& None
& $31.44 \pm 10.35$
& --
& 19.43
& 17.04
& 16.07 \\

& $\dot{q}_1$
& $34.96 \pm 9.73$
& $+11.20$
& 22.91
& 17.54
& 17.75 \\

& Altitude
& $33.80 \pm 7.67$
& $+7.52$
& 16.78
& 25.45
& 12.21 \\

& Throttle
& $\mathbf{30.31 \pm 11.27}$
& $-3.60$
& 17.80
& 18.83
& 14.68 \\
\hline
\end{tabular*}

\end{table*}

\begin{table*}
\centering
\caption{Robustness analysis on the unseen dataset over five random seeds.
Improvement is measured relative to the nominal linear strain FK.}
\label{tab:final_3d_performance}

\begin{tabular*}{\textwidth}{@{\extracolsep{\fill}}lccccc@{}}
\hline
\textbf{Model} &
\textbf{$x$ [mm]} &
\textbf{$y$ [mm]} &
\textbf{$z$ [mm]} &
\textbf{3D RMSE [mm]} &
\textbf{Improvement [\%]} \\
\hline

Linear strain FK
& 25.54
& 19.94
& 31.14
& 44.94
& -- \\

MLP
& $24.38 \pm 2.05$
& $\mathbf{13.26 \pm 2.68}$
& $23.24 \pm 4.23$
& $36.38 \pm 3.58$
& $19.04 \pm 7.98$ \\

GRU
& $15.58 \pm 2.22$
& $16.03 \pm 1.30$
& $16.31 \pm 2.23$
& $27.72 \pm 2.92$
& $38.32 \pm 6.50$ \\

CfC
& $\mathbf{11.92 \pm 0.85}$
& $14.27 \pm 1.86$
& $\mathbf{11.72 \pm 0.39}$
& $\mathbf{22.00 \pm 1.70}$
& $\mathbf{51.04 \pm 3.78}$ \\
\hline
\end{tabular*}

\end{table*}

Table~\ref{tab:final_3d_performance} presents the robustness analysis on the
unseen experiments. The nominal linear strain FK yields a 3D RMSE of
$44.94~\mathrm{mm}$. Across five random seeds, the MLP achieves
$36.38 \pm 3.58~\mathrm{mm}$, corresponding to a
$19.04 \pm 7.98\%$ improvement relative to the nominal model. The GRU reduces
the error to $27.72 \pm 2.92~\mathrm{mm}$, corresponding to an improvement
of $38.32 \pm 6.50\%$, while the CfC achieves
$22.00 \pm 1.70~\mathrm{mm}$ and an improvement of
$51.04 \pm 3.78\%$. In terms of mean 3D RMSE, the CfC therefore reduces the
error by approximately $39.5\%$ relative to the MLP and $20.6\%$ relative to
the GRU. It also exhibits the smallest variation in 3D RMSE across the five
random seeds and achieves the lowest 3D error in each run.

The component-wise results show that the CfC achieves the lowest RMSE in the
$x$ and $z$ directions, with $11.92 \pm 0.85~\mathrm{mm}$ and
$11.72 \pm 0.39~\mathrm{mm}$, respectively. Although the MLP yields the lowest
mean $y$-direction RMSE, its larger errors in the other directions result in a
substantially higher overall 3D RMSE, highlighting the importance of
evaluating the complete 3D position error rather than individual Cartesian
components. 

Performance also varies across the individual unseen experiments, with
Test~12 representing the most challenging condition for all three learned
estimators. This test exhibits relatively larger lateral CR deviation at close
ground proximity. The CfC yields lower errors than the GRU for Tests~11 and
12, whereas the two temporal models perform similarly on Test~10, for which
the GRU achieves a slightly lower mean error. Thus, the CfC does not
consistently outperform the GRU under every individual operating condition,
but provides lower overall error and greater consistency across the complete
unseen dataset.

Overall, the robustness analysis reinforces the benefit of temporal modeling
for UAV-induced residual estimation. Both temporal models achieve substantially
lower 3D RMSE than the memoryless MLP, while the CfC provides the lowest mean
3D RMSE and smallest variation under the experimental conditions
considered.

\section{Conclusion}

Two fundamental challenges in ACMs are investigated in this paper. These platforms are complex coupled systems with multibody and continuum dynamics, making accurate modeling challenging. Reduced-order models are therefore attractive for real-time deployment, although the trade-off between modeling accuracy and complexity must be carefully considered. Using an experimental dataset, a compact physics-informed strain model is evaluated and shown to provide accuracy comparable to that of a data-driven polynomial model while preserving a more structured representation. Real-world experiments also reveal noticeable deviations in CR end-effector position under UAV-induced aerodynamic effects. The effectiveness of learning-based residual estimation is therefore investigated, with temporal models providing substantially better performance than a memoryless approach on unseen experiments. These results demonstrate a lightweight and deployable approach for improving end-effector position estimation in ACMs. Future work will focus on extending the framework to more aggressive UAV maneuvers and stochastic aerodynamic conditions.

\bibliography{ref}

@IEEEtranBSTCTL{IEEEexample:BSTcontrol,
  CTLdash_repeated_names = "no"
}

@article{khamseh2018aerial,
  title={Aerial manipulation—A literature survey},
  author={Khamseh, Hossein Bonyan and Janabi-Sharifi, Farrokh and Abdessameud, Abdelkader},
  journal={Robotics and Autonomous Systems},
  volume={107},
  pages={221--235},
  year={2018},
  publisher={Elsevier}
}

@article{ollero2021past,
  title={Past, present, and future of aerial robotic manipulators},
  author={Ollero, Anibal and Tognon, Marco and Suarez, Alejandro and Lee, Dongjun and Franchi, Antonio},
  journal={IEEE Transactions on Robotics},
  volume={38},
  number={1},
  pages={626--645},
  year={2021},
  publisher={IEEE}
}

@article{jalali2022aerial,
  title={Aerial continuum manipulation: A new platform for compliant aerial manipulation},
  author={Jalali, Amir and Janabi-Sharifi, Farrokh},
  journal={Frontiers in Robotics and AI},
  volume={9},
  pages={903877},
  year={2022},
  publisher={Frontiers Media SA}
}

@article{amiri2025high,
  title={High-performance coupled kinematics of aerial continuum manipulation systems for control applications},
  author={Amiri, Niloufar and Janabi-Sharifi, Farrokh},
  journal={Robotics and Autonomous Systems},
  volume={192},
  pages={105021},
  year={2025},
  publisher={Elsevier}
}

@article{uthayasooriyan2026experimental,
  title={An Experimental Study of Downwash Effects on a Continuum Manipulator Integrated with a Multirotor UAV},
  author={Uthayasooriyan, Anuraj and Digumarti, Krishna Manaswi and Vanegas, Fernando and Gonzalez, Felipe},
  journal={IEEE Robotics and Automation Letters},
  year={2026},
  publisher={IEEE}
}

@article{liang2022adaptive,
  title={Adaptive prescribed performance control of unmanned aerial manipulator with disturbances},
  author={Liang, Jiacheng and Chen, Yanjie and Wu, Yangning and Miao, Zhiqiang and Zhang, Hui and Wang, Yaonan},
  journal={IEEE Transactions on Automation Science and Engineering},
  volume={20},
  number={3},
  pages={1804--1814},
  year={2022},
  publisher={IEEE}
}

@article{chen2022adaptive,
  title={Adaptive sliding-mode disturbance observer-based finite-time control for unmanned aerial manipulator with prescribed performance},
  author={Chen, Yanjie and Liang, Jiacheng and Wu, Yangning and Miao, Zhiqiang and Zhang, Hui and Wang, Yaonan},
  journal={IEEE transactions on cybernetics},
  volume={53},
  number={5},
  pages={3263--3276},
  year={2022},
  publisher={IEEE}
}

@article{liang2024observer,
  title={Observer-based nonlinear control for dual-arm aerial manipulator systems suffering from uncertain center of mass},
  author={Liang, Xiao and Wang, Yang and Yu, Hai and Zhang, Zhaopeng and Han, Jianda and Fang, Yongchun},
  journal={IEEE Transactions on Automation Science and Engineering},
  volume={22},
  pages={1984--1995},
  year={2024},
  publisher={IEEE}
}

@inproceedings{li2023nonlinear,
  title={Nonlinear mpc for quadrotors in close-proximity flight with neural network downwash prediction},
  author={Li, Jinjie and Han, Liang and Yu, Haoyang and Lin, Yuheng and Li, Qingdong and Ren, Zhang},
  booktitle={2023 62nd IEEE Conference on Decision and Control (CDC)},
  pages={2122--2128},
  year={2023},
  organization={IEEE}
}

@inproceedings{kharitenko2025spatiotemporal,
  title={A Spatiotemporal Downwash Modeling for Agile Close-Proximity Multirotor Flight},
  author={Kharitenko, Pavel and Fan, Yicheng and Liu, Xiaopei and Wang, Yang},
  booktitle={2025 IEEE/RSJ International Conference on Intelligent Robots and Systems (IROS)},
  pages={807--812},
  year={2025},
  organization={IEEE}
}

@article{wang2023neural,
  title={Neural moving horizon estimation for robust flight control},
  author={Wang, Bingheng and Ma, Zhengtian and Lai, Shupeng and Zhao, Lin},
  journal={IEEE Transactions on Robotics},
  volume={40},
  pages={639--659},
  year={2023},
  publisher={IEEE}
}

@article{bauersfeld2024robotics,
  title={Robotics meets fluid dynamics: A characterization of the induced airflow below a quadrotor as a turbulent jet},
  author={Bauersfeld, Leonard and Muller, Koen and Ziegler, Dominic and Coletti, Filippo and Scaramuzza, Davide},
  journal={IEEE Robotics and Automation Letters},
  volume={10},
  number={2},
  pages={1241--1248},
  year={2024},
  publisher={IEEE}
}

@article{chen2021adaptive,
  title={Adaptive modeling for downwash effects in multi-UAV path planning},
  author={Chen, Chih-Chun and Liu, Hugh H-T},
  journal={Guidance, Navigation and Control},
  volume={1},
  number={04},
  pages={2140005},
  year={2021},
  publisher={World Scientific}
}

@article{li2024adaptive,
  title={Adaptive neural network backstepping control method for aerial manipulator based on coupling disturbance compensation},
  author={Li, Hai and Li, Zhan and Liu, Jiayu and Zheng, Xiaolong and Yu, Xinghu and Kaynak, Okyay},
  journal={Journal of the Franklin Institute},
  volume={361},
  number={7},
  pages={106733},
  year={2024},
  publisher={Elsevier}
}

@article{fang2023robust,
  title={Robust control based on adaptive neural network for the process of steady formation of continuous contact force in unmanned aerial manipulator},
  author={Fang, Qian and Mao, Pengjun and Shen, Lirui and Wang, Jun},
  journal={Sensors},
  volume={23},
  number={2},
  pages={989},
  year={2023},
  publisher={MDPI}
}

@inproceedings{wu2024robust,
  title={Robust and energy-efficient control for multi-task aerial manipulation with automatic arm-switching},
  author={Wu, Ying and Zhou, Zida and Wei, Mingxin and Cheng, Hui},
  booktitle={2024 IEEE International Conference on Robotics and Automation (ICRA)},
  pages={8394--8400},
  year={2024},
  organization={IEEE}
}

@article{wang2023millimeter,
  title={Millimeter-level pick and peg-in-hole task achieved by aerial manipulator},
  author={Wang, Meng and Chen, Zeshuai and Guo, Kexin and Yu, Xiang and Zhang, Youmin and Guo, Lei and Wang, Wei},
  journal={IEEE Transactions on Robotics},
  volume={40},
  pages={1242--1260},
  year={2023},
  publisher={IEEE}
}

@article{zhang2026motion,
  author  = {Zhang, Yongzheng and Song, Hui and Hu, Zhaowen and Wei, Daozhu and Wang, Wei},
  title   = {Motion Control and Experimental Verification of a Continuum Aerial Manipulator for Power Grid Maintenance Operations},
  journal = {Journal of Field Robotics},
  pages   = {1--18},
  year    = {2026},
  doi     = {10.1002/rob.70276}
}

@article{amiri2026strain,
  title={Strain-Parameterized Coupled Dynamics and Dual-Camera Visual Servoing for Aerial Continuum Manipulators},
  author={Amiri, Niloufar and Janabi-Sharifi, Farrokh},
  journal={arXiv preprint arXiv:2603.23333},
  year={2026}
}

@article{boyer2020dynamics,
  title={Dynamics of continuum and soft robots: A strain parameterization based approach},
  author={Boyer, Frederic and Lebastard, Vincent and Candelier, Fabien and Renda, Federico},
  journal={IEEE transactions on robotics},
  volume={37},
  number={3},
  pages={847--863},
  year={2020},
  publisher={IEEE}
}

@article{renda2020geometric,
  title={A geometric variable-strain approach for static modeling of soft manipulators with tendon and fluidic actuation},
  author={Renda, Federico and Armanini, Costanza and Lebastard, Vincent and Candelier, Fabien and Boyer, Frederic},
  journal={IEEE Robotics and Automation Letters},
  volume={5},
  number={3},
  pages={4006--4013},
  year={2020},
  publisher={IEEE}
}

@article{lipton2015critical,
  author  = {Lipton, Zachary C. and Berkowitz, John and Elkan, Charles},
  title   = {A Critical Review of Recurrent Neural Networks for Sequence Learning},
  journal = {arXiv preprint arXiv:1506.00019},
  year    = {2015}
}

@article{hasani2022cfc,
  author  = {Hasani, Ramin and Lechner, Mathias and Amini, Alexander and
             Liebenwein, Lucas and Ray, Aaron and Tschaikowski, Max and
             Teschl, Gerald and Rus, Daniela},
  title   = {Closed-form continuous-time neural networks},
  journal = {Nature Machine Intelligence},
  volume  = {4},
  pages   = {992--1003},
  year    = {2022},
  doi     = {10.1038/s42256-022-00556-7}
}

@INPROCEEDINGS{11598604,
  author={Amiri, Niloufar and Sepahvand, Shayan and Mantegh, Iraj and Janabi-Sharifi, Farrokh},
  booktitle={2026 International Conference on Unmanned Aircraft Systems (ICUAS)}, 
  title={Systematic Analysis of Coupling Effects on Closed-Loop and Open-Loop Performance in Aerial Continuum Manipulators}, 
  year={2026},
  volume={},
  number={},
  pages={1184-1191},
  doi={10.1109/ICUAS69441.2026.11598604}}

@inproceedings{cho2014learning,
  title={Learning Phrase Representations using RNN Encoder--Decoder for Statistical Machine Translation},
  author={Cho, Kyunghyun and van Merrienboer, Bart and Gulcehre, Caglar and Bahdanau, Dzmitry and Bougares, Fethi and Schwenk, Holger and Bengio, Yoshua},
  booktitle={Proceedings of the 2014 Conference on Empirical Methods in Natural Language Processing (EMNLP)},
  pages={1724--1734},
  year={2014}
}

\bibliographystyle{IEEEtran}

\end{document}